\pdfoutput=1
\documentclass[11pt]{article}

\usepackage[preprint]{acl}

\usepackage{times}
\usepackage{latexsym}
\usepackage{makecell}

\usepackage[T1]{fontenc}
\usepackage[utf8]{inputenc}

\usepackage{microtype}

\usepackage{inconsolata}

\usepackage{graphicx}

\usepackage{todonotes}
\usepackage{xspace}
\definecolor{ourred}{HTML}{E13342}
\definecolor{ourblue}{HTML}{6495ed}
\definecolor{Gray}{gray}{0.92}
\definecolor{racing-green}{rgb}{0.0, 0.8, 0.6}
\definecolor{awesome-red}{rgb}{1.0, 0.13, 0.32}
\definecolor{lgrey}{HTML}{787878}
\definecolor{mgrey}{HTML}{656565}
\definecolor{hgrey}{HTML}{9B9B9B}
\definecolor{black}{HTML}{000000}

\usepackage{xcolor, colortbl}

\usepackage{microtype}
\usepackage{amsmath}
\usepackage{amsfonts} 
\usepackage{algorithm}
\usepackage{algorithmic}
\usepackage{graphicx}
\usepackage{subfigure}
\usepackage{tabularray}
\usepackage{wrapfig}
\usepackage{float}
\usepackage{caption}

\usepackage{pifont}
\definecolor{racing-green}{rgb}{0.0, 0.8, 0.6}
\definecolor{awesome-red}{rgb}{1.0, 0.13, 0.32}
\definecolor{gblue}{HTML}{787878}
\definecolor{gorange}{HTML}{656565}
\definecolor{gred}{HTML}{9B9B9B}
\definecolor{ouryellow}{HTML}{ECCF65}
\definecolor{ourgrey}{HTML}{483249}

\usepackage{newunicodechar}
\usepackage{enumitem}

\usepackage{hyperref}
\usepackage{url}
\usepackage{booktabs}
\usepackage{multirow}
\usepackage{hhline}
\usepackage{inconsolata}
\usepackage{amsmath}
\usepackage{wrapfig}

\usepackage{amssymb}
\usepackage{microtype}
\usepackage{listings}
\lstdefinestyle{text_rm}{
    basicstyle=\ttfamily\small,
    commentstyle=\color{gray},
    stringstyle=\color{blue},
    keywordstyle=\color{red},
    numbers=none,
    breakindent=0pt,
    frame=none,
    breaklines=true,
    language=TeX,
    moredelim=**[is][\bfseries]{@}{@}
}

\makeatletter
\def\hlinewd#1{%
\noalign{\ifnum0=`}\fi\hrule \@height #1 %
\futurelet\reserved@a\@xhline}
\makeatother

\definecolor{darkblue}{rgb}{0, 0, 0.5}
\hypersetup{colorlinks=true, citecolor=darkblue, linkcolor=darkblue, urlcolor=darkblue}

\usepackage{algorithm}
\title{Evaluating LLM-based Approaches to Legal Citation Prediction: Domain-specific Pre-training, Fine-tuning, or RAG? \\A Benchmark and an Australian Law Case Study}
\author{
  Jiuzhou Han$^{{\natural} }$\ \ \ \ \ 
  Paul Burgess$^{{\sharp}}$\ \ \ \ \ 
  Ehsan Shareghi$^{{\natural} }$\\
  $^{{\natural} }$~Department of Data Science \& AI, Monash University \\
  $^{{\sharp}}$~Faculty of Law, Monash University\\
  \texttt{first.last@monash.edu}
}

\begin{document}
\maketitle
\begin{abstract}
Large Language Models (LLMs) have demonstrated strong potential across legal tasks, yet the problem of \emph{legal citation prediction} remains under-explored. At its core, this task demands fine-grained contextual understanding and precise identification of relevant legislation or precedent. We introduce the \emph{AusLaw Citation Benchmark}, a real-world dataset comprising $55k$ Australian legal instances and $18,677$ unique citations which to the best of our knowledge is the first of its scale and scope. We then conduct a systematic benchmarking across a range of solutions: (i) standard prompting of both general and law-specialised LLMs, (ii) retrieval-only pipelines with both generic and domain-specific embeddings, (iii) supervised fine-tuning, and (iv) several hybrid strategies that combine LLMs with retrieval augmentation through query expansion, voting ensembles, or re-ranking. Results show that neither general nor law-specific LLMs suffice as stand-alone solutions, with performance near zero. Instruction tuning (of even a generic open-source LLM) on task-specific dataset is among the best performing solutions. We highlight that database granularity along with the type of embeddings play a critical role in retrieval-based approaches, with hybrid methods which utilise a trained re-ranker delivering the best results. 
Despite this, a performance gap of nearly $50\%$ remains, underscoring the value of this challenging benchmark as a rigorous test-bed for future research in legal-domain.\footnote{The data, code, and trained LLMs and re-rankers are available at:  \url{https://auslawbench.github.io}}
%\footnote{For code and data, see \url{https://anonymous.4open.science/r/LegalCitationPrediction-5606}. Models will be available with publication.}
\end{abstract}

\section{Introduction}\label{sec:introduction}
%\begin{itemize}
Recent advancements in utilising Large Language Models (LLMs) for legal domain have shown promising results across various tasks. For instance, \citet{pont2023legal} leveraged LLMs for generating summaries of judicial decisions, identifying legal issues, decision-making criteria, and specifying keywords. \citet{deroy2024applicability} revealed that LLMs outperform extractive summarisation methods in quality metrics but suffer from inconsistencies and hallucinations, highlighting the importance of human-in-the-loop approaches for improved reliability. \citet{10.1145/3594536.3595170} underscored that fine-tuning LLMs demonstrates state-of-the-art performance in legal judgment prediction, while~\citet{peng2024athena} showed retrieval-augmentation leads to improved accuracy by integrating external knowledge, particularly for complex charges. \citet{hou2024gaps} proposed a model to detect deviations between an AI-generated legal analysis and human as a way of quantifying their reliabilities.  

While these advancements highlight LLMs’ transformative potential in legal applications, challenges like ensuring factual accuracy, handling diverse tasks, and mitigating inherent issues such as hallucination remain. For instance, \citet{dahl2024large} reports that even state-of-the-art LLMs hallucinate between 69-88\% of responses to legal queries, while \citet{magesh2024hallucination} highlights that hallucination issue is mitigated with specialisation of tools but still persists as an unresolved issue. 

%While we provide a brief overview of the literature in this paper (\S\ref{sec:background}), a comprehensive overview of LLMs application in various areas of legal domain is beyond the scope of this paper . 

In this paper, we report progress with respect to a less-explored  task in the legal domain, \emph{Legal Citation Prediction}. Citations in legal cases, as in academic writing, serve two purposes: first, information necessary to locate, read, and verify the material; and second, information about the authority of the source is conveyed~\citep{AxelLute1982}. The second is particularly crucial in legal cases in common law jurisdictions, as in those systems most decision makers are required to follow previous decisions~\citep{Schauer1987}. Accordingly, whilst precedent plays a fundamental role in determining how courts behave and, therefore, in how societies function, citations to, of, and between authorities are the way in which precedent is communicated. In determining matters in court, judges use citations to give weight and authority to their decisions and also, crucially, to demonstrate that they are acting appropriately and are adhering to the legal precedent that already exists. When doing so, a judge or a panel of judges will often state a legal rule -- as a principal or a proposition -- and then support the existence of that rule with a citation to another source; frequently a prior court decision. In other words, judges making determinations today rely on citations to not only determine legal questions in court but also to show that they are acting within their (the judges') lawful role.  

Formally, the citation prediction task can be defined as follows: Given a passage, the goal is to identify the correct legislation or precedent that applies and needs to be cited (i.e., to predict or retrieve the correct \emph{[CITATION]}). The following examples illustrate the citation prediction tasks considered in this paper:
\begin{description}[noitemsep,leftmargin=2.5mm,topsep=0pt,parsep=0pt,partopsep=0pt]
    \item[\textsc{Example 1.}] \underline{Query}: \emph{The distinction between a genuine offer of compromise and a demand to capitulate has to be recognised. See the discussion in [CITATION].\newline \underline{Answer}: Leichhardt Municipal Council v Green [2004] NSWCA 341}
    \item[\textsc{Example 2.}] \underline{Query}: \emph{The Tribunal is satisfied that the applicant does not fulfil the requirements of section 139(a) of the National Law, in that she lacks the mental capacity to practise medicine, as was considered in [CITATION]. \newline \underline{Answer}: Lindsay v Health Care Complaints Commission [2010] NSWCA 194}
    \item[\textsc{Example 3.}] \underline{Query}: \emph{Whilst it is suggested that the offender’s mother and grandmother have difficulty paying rent without the offender’s assistance, there is no evidence of how they support themselves or their financial circumstances. There is no evidence of hardship that might meet the ‘truly, wholly or highly exceptional’ standard referred to in [CITATION]. \newline \underline{Answer}: Jinnette v R [2012] NSWCCA 217}
\end{description}

We aim to compare, develop, and explore different solutions for the citation prediction task. We release, \textbf{AusLaw Citation Benchmark}, a real dataset of $55k$ instances specific to Australian law, covering $18,677$ unique citations. To the best of our knowledge this is the first dataset of this scale and scope. We then conduct a thorough comparison along the following comprehensive dimensions: 
\begin{itemize}[noitemsep,leftmargin=2.5mm,topsep=0pt,parsep=0pt,partopsep=0pt]
    \item Prompting general purpose LLMs (i.e., GPT-4o~\citep{gpt4o,gpt4}, Claude Sonnet 3.5~\citep{anthropic2024claude35sonnet}, LLaMA-3.1-70B-instruct~\citep{grattafiori2024llama3}, Command R+~\citep{commandrplus}) 
    \item Prompting law-specialised pre-trained LLMs (i.e., SaulLM-7B-instruct~\citep{saul7b}, SaulLM-54B-instruct~\citep{saul54b})
    \item Retrieval-only setup with vectorised DB using general-purpose (i.e., text-embedding-3-large) and law-specific embeddings (i.e., AusLaw-embedding-v1.0)
    \item Instruction fine-tuning LLMs (i.e., SaulLM-7B, and LLaMA-3.1-8B) for the citation prediction task
    \item Different hybrid tactics that combine LLMs and retrieval systems (i.e., retrieval-augmented generation, query expansion, voting ensemble, and specialised re-rankers)
\end{itemize}
While we aim to push all these solutions to their limits to better understand their effectiveness and limitations, a near $50\%$ gap is remained to be filled. We hope this benchmark and the baselines and methods investigated in this paper to encourage future developments in this important and challenging task.

%\textcolor{red}{TBD: In doing this, we highlight that general purpose LLMs as stand-alone tools are highly incapable, with specialised pre-trained LLMs being slightly better at the task. ...}

% Will add this in the 2nd version - for now parking this
\section{Legal NLP in the Era of LLMs}\label{sec:background}
While a satisfying review demands a separate work, we attempt to provide a brief overview of existing work on the intersection of Law and LLMs. We also acknowledge that this only provides an overview of a small subset of the broader space of research in Legal NLP. 

The intersection of NLP and the legal domain has given rise to a wide array of research tackling tasks such as legal text classification, retrieval, summarisation, question answering, and reasoning. As legal texts are often complex and domain-specific, specialised benchmarks and models have become essential to evaluate and improve NLP performance in this high-stakes field.

\subsection{Legal Benchmarks}
Recent efforts in benchmarking legal NLP have produced a diverse ecosystem of datasets targeting legal reasoning, retrieval, and comprehension. Benchmarks like LEGALBENCH~\cite{guha2024legalbench}, LawBench~\cite{fei2023lawbench}, and LexEval offer broad, multi-task evaluations across cognitive levels and languages, revealing persistent limitations in current LLMs’ legal reasoning abilities. Task-specific benchmarks like BLT~\cite{blair2023blt} and MAUD~\cite{wang2023maud} assess fundamental skills like basic legal text navigation and merger agreement comprehension, while CUAD~\cite{hendrycks2021cuad} and the statutory reasoning~\cite{DBLP:journals/corr/abs-2005-05257} in tax law dataset bring attention to contract review and rule-based reasoning tasks demanding deep understanding of legal semantics and structure. LexGLUE~\cite{chalkidis-etal-2022-lexglue} and LEGAL-BERT~\cite{legalbert} contribute to general legal language understanding and adaptation of pretrained models for legal tasks, while LegalBench-RAG~\cite{legal-rag} and CLERC~\cite{abe2024clerc} evaluate retrieval and generation in legal writing tasks. Newer benchmarks such as LegalAgentBench~\cite{li2024legalagentbench} assess autonomous legal agents through multi-step tasks and tool use in real-world scenarios, while LegalHalBench~\cite{hu2025fine} introduces fine-grained metrics and datasets to evaluate and reduce hallucination in legal QA systems. Region-specific datasets like IL-TUR~\cite{indian-bench} for Indian law and LexEval~\cite{chinese-bench} for Chinese legal systems underscore the need for jurisdictionally relevant evaluations. We propose a new benchmark focused on legal citation prediction. 

\subsection{Law-Specialised Large Language Models}
The development of law-specialised LLMs has gained momentum in response to the distinctive demands of legal language, which is domain-specific, syntactically complex, and semantically nuanced. Early adaptation efforts such as LEGAL-BERT~\cite{legalbert} explored various strategies for pretraining BERT models on legal corpora, revealing that full domain-specific pretraining outperforms generic approaches. More recent models like Lawma~\cite{lawma} and LawLLM~\cite{lawllm} demonstrate the effectiveness of fine-tuned and multi-task architectures in addressing legal classification, retrieval, and judgment prediction tasks within the United States legal system. The SaulLM~\cite{saul7b, saul54b} family is the first (and only) open-source Law-specialised LLMs ranging from 7B to 141B parameters, using extensive legal corpora (500B tokens), instruction tuning, and preference alignment to achieve state-of-the-art results across multiple legal benchmarks. These models highlight the benefits of legal specialisation. 

Nevertheless, the presence of domain-specialised LLMs only reduce the semantic gap required for the domain and still require (as we will show later in the experiments) task-specific training to be competent for fine-grained tasks such as legal citation prediction. Additionally, we argue (and empirically show) that pretraining on a mixture of data from various jurisdictions and countries does not provide a reliable understanding of the legal context specific to each of those jurisdictions. To bridge this gap, we develop an Australia-specific legal LLM focused on tasks such as legal citation prediction—thereby supporting jurisdiction-aware applications and advancing the broader field of legal artificial intelligence.

% \subsection{AI Research for Law in Australia}

\section{AusLaw Citation Benchmark}\label{sec:data}
To build our benchmark, we use the NSW Caselaw section of the Open Australian Legal Corpus.\footnote{\url{https://huggingface.co/datasets/umarbutler/open-australian-legal-corpus}} We identified $82,530$ citations that specifically referred to a case within the dataset. For each citation ($c_k$), we extracted the sentence in which the citation appeared (denoted as $S^i_{c_k}$), along with the preceding sentence (denoted as $S^{i-1}_{c_k}$). We further utilised an LLM to generate an auxiliary description of $c_k$, based on (${{Full\  Text}_{c_k}}, S^i_{c_k},S^{i-1}_{c_k}$) where ${Full\  Text}_{c_k}$ refers to the full text of the cited case. We manually checked a subset of the LLM-generated descriptions for quality assurance and optimizing the prompt wording. An example of a generated \emph{RoC} is provided below:
\begin{quote}
    \underline{\textbf{$S^{i-1}_{c_k},S^{i}_{c_k}$:}} \emph{In considering what if any orders should be made in regards to the surface roots, I am not satisfied to the level required by s 10(2) of the Act, that there is any real likelihood of injury arising from those roots. Craig J in Leichhardt Municipal Council v Green [2004] NSWCA 341, considered that 'something more than a theoretical possibility is required in order to engage the power under the Trees Act'. \newline\underline{CITATION $c_k$:} Leichhardt Municipal Council v Green [2004] NSWCA 341\newline\underline{Generated Reason-of-Citation (RoC)}: The cited case is referenced to establish the standard required to demonstrate a likelihood of injury under the Trees Act.}
\end{quote}

We refer to this description as  \emph{Reason-of-Citation} (RoC).\footnote{In this process, we tasked the LLM to not generate any {RoC} (i.e., to generate \texttt{NOT ENOUGH INFORMATION}) if the information was not adequate in $(S^i_{c_k},S^{i-1}_{c_k})$. These refused cases were not included in the data. See Table~\ref{prompt:roc} of Appendix~\ref{app:prompt} for details on when this was triggered.} Each reference to a citation $c_k$ in the data results in a unique new $\text{RoC}_{c_k}$. For each \( c_k \), we denote its \( M \) references as \(\mathrm{RoC}_{c_k}^1, \mathrm{RoC}_{c_k}^2, \ldots, \mathrm{RoC}_{c_k}^M\). We will discuss later how the {RoC}s are used. This resulted in the final dataset of $55,005$ instances, covering $18,677$ unique citations. Within this set, 5\% of the citations have been referenced at least 9 times, while 54\% were cited only once. From this final set, we extracted $1k$ citations as test set, and used the rest for training (i.e., our instruction-tuned models). See Appendix~\ref{sec:visual} for more details.

\section{Methods}\label{sec:method}
We investigate various methods under Open World and Closed World settings. The Open World setting places no restriction over the possible predictions from the system (i.e., similar to how an LLM functions in real-world), whereas the Closed World setting confines the output space to be from the set of $18,677$ citations present in the database (i.e., similar to a standard retrieval setup). 

\subsection{LLM-Only}\label{subsec:llm}
We explored both existing LLMs (General purpose and Law-specialised) as well as our instruction-tuned LLMs. 
\paragraph{Existing General Purpose LLMs.} For the general purpose LLMs we used GPT-4o~\citep{gpt4o,gpt4}, Claude Sonnet 3.5~\citep{anthropic2024claude35sonnet}, LLaMA-3.1-70B-instruct~\citep{grattafiori2024llama3}, and Command R+~\citep{commandrplus}. When directly prompted with the query, all these LLMs demonstrated near-zero performance. To gain deeper insight, we  leaked the \emph{RoC}s along with the query text. This approach, while bypassing the task's requirement of predicting the citation without access to any information about the case-to-be-cited, allowed us to estimate the upper-bound performance of these models. More on this will be discussed in Section~\ref{sec:experiment}.
\paragraph{Existing Law-specialised LLMs.}For law-specialised LLMs, we used SaulLM-7B-instruct~\citep{saul7b}, SaulLM-54B-instruct~\citep{saul54b} which to the best of our knowledge are the only publicly available law LLMs for English to this date.\footnote{We were unable to run SaulLM-141B-Instruct due to its substantial hardware requirements.} For prompting, we followed the exact setting of general purpose LLM experiments (see Table~\ref{prompt:llmonly} of Appendix~\ref{app:prompt}).

\paragraph{Our Instruction Fine-tuned LLMs.} 
We instruction-tuned LLaMA-3.1-8B and SaulLM-7B-Base models on the training data. These resulting models are referred to as Cite-LLaMA-3.1-8B and Cite-SaulLM-7B in our experiments. At inference time, given the query ($S^{i-1}_{c_k}, \texttt{Mask}(S^i_{c_k})$) the model first produces the $\texttt{RoC}_{c_k}$ and then $c_k$ (i.e., $p(c_k,\texttt{RoC}_{c_k} \mid I, S^{i-1}_{c_k}, \texttt{Mask}(S^i_{c_k}); \theta)$:
\begin{align}
&p(c_k \mid I,S^{i-1}_{c_k},\texttt{Mask}(S^{i}_{c_k}),\texttt{RoC}_{c_k};\theta)\\\nonumber
   &\quad\times p(\texttt{RoC}_{c_k}\mid I,S^{i-1}_{c_k},\texttt{Mask}(S^{i}_{c_k});\theta).
\end{align}
where $\theta$ denotes LLM parameters, $I$ denotes the instruction, $\texttt{Mask}(S^i_{c_k})$ denotes $S^i_{c_k}$ with the citation to $c_k$ being masked. See Appendix~\ref{app:instructiontuning} for further details on training parameters and instruction detail and format (same instruction was used at inference).

Unlike the previous setups, for all experiments with the instruction-tuned models only the text of the query (i.e., no {RoC}) was provided and the model is instructed to predict both the $RoC_{c_k}$ and the citation $c_k$. This presents a challenging setup as the LLM needs to predict a citation in the Open World setting solely by its parametric knowledge and the brief information provided in the query text. 

\subsection{Retrieval-Only} We investigated retrieval along two axes: embeddings, and granularity of data to be indexed in the vectorised database. The basic principle is a retrieval task where a query is matched against all records of a database, and the Top-k closest records are returned. The closeness is measured in the semantic space through embeddings. In all  setups involving retrieval, unless stated otherwise, the search returns the Top-5 most relevant citations.

\paragraph{Embeddings.} For embeddings in the retrieval system (both for representation of queries and the vectorized database) the 3072 dimensional \texttt{text-embedding-3-large}\footnote{\url{https://openai.com/index/new-embedding-models-and-api-updates}} from OpenAI as generic embeddings, and the 384 dimensional vectors of \texttt{AusLaw-embedding-v1.0}~\footnote{\url{https://huggingface.co/adlumal/auslaw-embed-v1.0}} as Australian law-specialised embeddings were used.\footnote{While we used the largest dimensionality of embeddings available to measure the performance upper-bounds, there are directions of exploration which we did not pursue in this work: the impact of using various dimensionalities on the retrieval system, or alternative means of building dense or conventional types of indexes (i.e., Lucene/BM25).}

\paragraph{Database Granularities.} To construct the database, three possibilities for representing each citation ($c_k$) were considered: (1) Full Case Text ($\text{Full Case}_{c_k}$), (2) Catchwords~($\text{Catchwords}_{c_k}$), and (3) Aggregation of all its corresponding $M$ {RoC}s ($\text{RoC Aggregations} = \operatorname{concat}(\mathrm{RoC}^1_{c_k}, \mathrm{RoC}^2_{c_k}, \ldots, \mathrm{RoC}^M_{c_k})$). 

We refer to these different settings as Index Granularity later in the experiments (i.e., Table~\ref{tab:main}). For each granularity level, we created a distinct database version for each embedding backbone.\footnote{We use \url{https://github.com/chroma-core/chroma} for building the DB, and use the built-in cosine similarity for measuring closeness.}

\subsection{Hybrid of LLM and Retrieval}
What follows can be regarded as various instances of Retrieval Augmented Generation (RAG). However, in our manuscript, we strictly refer to RAG as the scenario where the LLM generation is guided by retrieval, rather than retrieval being influenced by the generation. 

\paragraph{Query Expansion.} In this setting, given the query, the LLM was first asked to produce a potential description of a good citation. We denote this as RoC$^\text{aux}$ to underscore our deliberation in eliciting what the LLM could semantically generate as an auxiliary RoC. The query is then expanded from Text to Text+RoC$^\text{aux}$. The expanded query is launched into the database, following the standard retrieval setup. See {Table~\ref{prompt:llm-retrieval}} of Appendix~\ref{app:prompt} for prompt details).

\paragraph{Voting Ensemble} First, following the LLM-only setup, instruction-tuned LLMs (e.g., Cite-LLaMA-3.1-8B) generate the RoC' and the citation. Next, the query is expanded into Text+RoC' and fed into the retrieval system. If the LLM-generated citation appears in the Top-5 retrieval results, it is returned; otherwise, the Top-1 result from the retrieval system is returned. This is to leverage the benefit of LLM-only setup by reducing its hallucination (i.e., producing citations outside the corpus).

\paragraph{Retrieval Augmented Generation (RAG).} This follows a standard Retrieval-Augmented Generation (RAG) setup, where queries are sent to the database, and the Top-5 results are retrieved for the LLM to re-rank and select the Top-1 match. See Table~\ref{prompt:llm-ranking} of Appendix~\ref{app:prompt} for prompt details.

\paragraph{Re-ranker.} We have four settings to train the re-ranker. The same data used for instruction fine-tuning was used for training the re-rankers. The input of a re-ranker is the text containing the missing citation and retrieved top-5 candidate citations and their corresponding RoC. The output is the gold citation. Since the RoC of each case has different length, to train a re-ranker, we needed to eliminate the effect of discrepancy in the length. 
\begin{itemize}
[noitemsep,leftmargin=3mm,topsep=0pt,parsep=0pt,partopsep=0pt]
    \item Setting 1 uses the first reason as the whole citation reason. 
    \item Setting 2 tasks GPT-4o-mini to merge the citation reasons of each case into a similar length text. 
    \item Setting 3 leverages the same merged citation reason as Setting 2. In addition, given 5 cases and their case texts, the GPT-4o-mini is tasked to generate the rational of why the gold one is cited among all 5 citations. 
    \item Setting 4 is based on retrieval. We retrieve the most similar citation reason from the citation reasons list as the citation reason of that case. Within this setting, there are two variations: 1) we use the retrieved top-1 citation reason as the output for the training; 2) we use the gold citation reason as the output for the training. We name these two variations as 4v1 and 4v2.
\end{itemize}
See {Table~\ref{reranker_instructions}} of Appendix~\ref{app:instructiontuning} for input and output details.

\section{Experiments}\label{sec:experiment}

\begin{table*}[!htbp]
\centering
\setlength{\extrarowheight}{2pt}
\renewcommand{\arraystretch}{0.9}
\setlength{\tabcolsep}{7pt}
\resizebox{\linewidth}{!}{%
\begin{tabular}{lllcccc}
\hlinewd{1pt}
\hline
\rowcolor{gray!20}
\multicolumn{7}{c}{\textbf{LLM-only Approach: Direct zero-shot prompting}}\\
\rowcolor{gray!20}
&Type&LLMs&Query &Output&ACC@1&ACC@5\\
\parbox[t]{2mm}{\multirow{8}{*}{\rotatebox[origin=c]{90}{Open World}}} &\multirow{4}{*}{General Purpose}&GPT-4o&Text+RoC&Top-5 Citations&0.1&0.1\\
&&Claude Sonnet 3.5&Text+RoC&Top-5 Citations&15.5&16.8\\
&&Command R+&Text+RoC&Top-5 Citations&0.0&0.0\\
&&LLaMA 3.1 70B Instruct&Text+RoC&Top-5 Citations&1.6&2.1\\
\cline{3-7} 
&\multirow{2}{*}{Law-specialised}&SaulLM-7B-Instruct&Text+RoC&Top-5 Citations&0.0&0.0\\
&&SaulLM-54B-Instruct&Text+RoC&Top-5 Citations&2.0&2.7\\
\cline{3-7} 
&\multirow{2}{*}{Citation-tuned (ours)} &\cellcolor{green!25}Cite-SaulLM-7B&\cellcolor{green!25}Text&\cellcolor{green!25}RoC+Top-1 Citation&\cellcolor{green!25}51.7$^\ast$&\cellcolor{green!25}-\\
&&\cellcolor{green!25}Cite-LLaMA-3.1-8B&\cellcolor{green!25}Text&\cellcolor{green!25}RoC+Top-1 Citation&\cellcolor{green!25}46.2&\cellcolor{green!25}-\\
\hlinewd{1pt}
\rowcolor{gray!20}
\multicolumn{7}{c}{\textbf{Retrieval-only Approach: Uses vectorised database and vectorised Query to retrieve Top-5}}\\
\rowcolor{gray!20}
\parbox[t]{2mm}{\multirow{8}{*}{\rotatebox[origin=c]{90}{}}}&Embeddings&Index Granularity& Query &Output&ACC@1&ACC@5\\
&\multirow{3}{*}{text-embedding-3-large}&Full Cases&Text&Top-5 Citations&14.9&32\\
%&&Full Cases&Text+RoC&Top-5 Citations&19.4&37.2\\
%\cline{3-7} 
&&Catchwords&Text&Top-5 Citations&14.7&32.5\\
%&&Catchwords&Text+RoC&Top-5 Citations&15.7&33.6\\
%\cline{3-7}
&&\cellcolor{red!25}RoC Aggregations&\cellcolor{red!25}Text&\cellcolor{red!25}Top-5 Citations&\cellcolor{red!25}27.1&\cellcolor{red!25}53.8\\
%&&RoC Aggregations&Text+RoC&Top-5 Citations&42.7&66.3\\
\cline{3-7} 
&\multirow{3}{*}{AusLaw-embedding}&Full Cases&Text&Top-5 Citations&8.7&20.7\\
%&&Full Cases&Text+RoC&Top-5 Citations&13.6&25.9\\
%\cline{3-7} 
&&Catchwords&Text&Top-5 Citations&10.5&22.4\\
%&&Catchwords&Text+RoC&Top-5 Citations&10.5&24.0\\
%\cline{3-7} 
&&\cellcolor{orange!25}RoC Aggregations&\cellcolor{orange!25}Text&\cellcolor{orange!25}Top-5 Citations&\cellcolor{orange!25}29.5&\cellcolor{orange!25}54.5\\
%&&RoC Aggregations&Text+RoC&Top-5 Citations&36.8&57.9\\
\hlinewd{1pt}

\rowcolor{gray!20}
\multicolumn{7}{c}{\textbf{ (Hybrid Approach) Query Expansion: Given Query, RoC$^\text{aux}$ is generated by an LLM and Query+RoC$^\text{aux}$ is used for retrieval}}\\
\rowcolor{gray!20}
\multicolumn{7}{c}{\textbf{Results are formatted as GPT-4o/SaulLM-54B-Instruct/Cite-LLaMA-3.1-8B/Cite-SaulLM-7B}}\\
\rowcolor{gray!20}
&Embeddings&Index Granularity& Query &Output&ACC@1&ACC@5\\
\parbox[t]{2mm}{\multirow{6}{*}{\rotatebox[origin=c]{90}{}}}&\multirow{3}{*}{text-embedding-3-large}&Full Cases&Text&Top-5 Citations&14.3/14.4/17.1/17.4&31.1/31.4/34.3/34.4\\
&&Catchwords&Text&Top-5 Citations&15.3/15.5/15.0/15.8&33.1/33.1/33.4/33.9\\
&&\cellcolor{red!25}RoC Aggregations&\cellcolor{red!25}Text&\cellcolor{red!25}Top-5 Citations&\cellcolor{red!25}29.6/28.6/34.9/35.1&\cellcolor{red!25}56.7/56.1/60.0/60.4\\
\cline{2-7} 
&\multirow{3}{*}{AusLaw-embedding}&Full Cases&Text&Top-5 Citations&9.0/9.5/11.7/12.4&21.1/21.3/24.2/26.0\\
&&Catchwords&Text&Top-5 Citations&10.2/10.9/11.0/11.4&23.5/24.6/24.3/24.4\\
&&\cellcolor{orange!25}RoC Aggregations&\cellcolor{orange!25}Text&\cellcolor{orange!25}Top-5 Citations&\cellcolor{orange!25}32.2/30.4/33.5/34.7&\cellcolor{orange!25}55.8/54.2/55.6/56.5\\
\hlinewd{1pt}

\rowcolor{gray!20}
\multicolumn{7}{c}{\textbf{(Hybrid Approach) Voting Ensemble: Returns LLM's citation if in the Top-5 of retrieval; otherwise, returns the retrieval's Top-1}}\\
\rowcolor{gray!20}
\multicolumn{7}{c}{\textbf{Results are formatted as Cite-LLaMA-3.1-8B/Cite-SaulLM-7B}}\\
\rowcolor{gray!20}
&Embeddings&Index Granularity& Query &Output&ACC@1&ACC@5\\
&text-embedding-3-large&\cellcolor{red!25}RoC Aggregations&\cellcolor{red!25}Text&\cellcolor{red!25}Top-5 Citations&\cellcolor{red!25}47.3/48.2&\cellcolor{red!25}-\\
&AusLaw-embedding&\cellcolor{orange!25}RoC Aggregations&\cellcolor{orange!25}Text&\cellcolor{orange!25}Top-5 Citations&\cellcolor{orange!25}43.6/45.3&\cellcolor{orange!25}-\\
\hlinewd{1pt}

\rowcolor{gray!20}
\multicolumn{7}{c}{\textbf{(Hybrid Approach) RAG: Given the Query, first retrieves Top-5 citations, and uses GPT-4o to pick the best}}\\
\rowcolor{gray!20}
\parbox[t]{2mm}{\multirow{8}{*}{\rotatebox[origin=c]{90}{}}}&Embeddings&Index Granularity& Query &Output&ACC@1&ACC@5\\
&\multirow{3}{*}{text-embedding-3-large}&Full Cases&Text&Top-5 Citations&16.5$^\ddagger$&-\\
&&Catchwords&Text&Top-5 Citations&21.7&-\\
&&\cellcolor{red!25}RoC Aggregations&\cellcolor{red!25}Text&\cellcolor{red!25}Top-5 Citations&\cellcolor{red!25}42.2&\cellcolor{red!25}-\\
\cline{2-7} 
&\multirow{3}{*}{AusLaw-embedding}&Full Cases&Text&Top-5 Citations&10.2$^\ddagger$&-\\
&&Catchwords&Text&Top-5 Citations&17.1&-\\
&&\cellcolor{orange!25}RoC Aggregations&\cellcolor{orange!25}Text&\cellcolor{orange!25}Top-5 Citations&\cellcolor{orange!25}42.9&\cellcolor{orange!25}-\\
\hlinewd{1pt}
\rowcolor{gray!20}
\multicolumn{7}{c}{\textbf{(Hybrid Approach) Re-ranker: Given the Query, first retrieves Top-5 citations, and use trained re-ranker to pick the best}}\\
\rowcolor{gray!20}
\multicolumn{7}{c}{\textbf{Results are formatted as setting1/2/3/4v1/4v2}}\\
\rowcolor{gray!20}
%\parbox[t]{2mm}{\multirow{8}{*}{\rotatebox[origin=c]{90}{Closed World}}}
&Embeddings&Index Granularity& Query &Output&ACC@1&ACC@5\\
%&\multirow{3}{*}{text-embedding-3-large}&Full Cases&Text&Top-5 Citations&&-\\
%&&Catchwords&Text&Top-5 Citations&&-\\
&text-embedding-3-large&\cellcolor{red!25}RoC Aggregations&\cellcolor{red!25}Text&\cellcolor{red!25}Top-5 Citations&\cellcolor{red!25}20.2/26.9/21.7/50.4/50.9&\cellcolor{red!25}-\\
\cline{2-7} 
%&\multirow{3}{*}{AusLaw-embedding}&Full Cases&Text&Top-5 Citations&&-\\
%&&Catchwords&Text&Top-5 Citations&&-\\
&AusLaw-embedding&\cellcolor{orange!25}RoC Aggregations&\cellcolor{orange!25}Text&\cellcolor{orange!25}Top-5 Citations&\cellcolor{orange!25}20.6/28.5/22.9/50.2/51.2&\cellcolor{orange!25}-\\
\hlinewd{1pt}
\rowcolor{gray!20}
\multicolumn{7}{c}{\textbf{(Hybrid Approach) Re-ranker: Given the Query, RoC$^\text{aux}$ is generated by Cite-SaulLM-7B.}}\\
\rowcolor{gray!20}
\multicolumn{7}{c}{\textbf{The Query+RoC$^\text{aux}$ is used for top-5 retrieval and passed to trained re-ranker to pick the best.}}\\
\rowcolor{gray!20}
\multicolumn{7}{c}{\textbf{Results are formatted as setting1/2/3/4v1/4v2}}\\
\rowcolor{gray!20}
%\parbox[t]{2mm}{\multirow{8}{*}{\rotatebox[origin=c]{90}{Closed World}}}
&Embeddings&Index Granularity& Query &Output&ACC@1&ACC@5\\
%&\multirow{3}{*}{text-embedding-3-large}&Full Cases&Text&Top-5 Citations&&-\\
%&&Catchwords&Text&Top-5 Citations&&-\\
&text-embedding-3-large&\cellcolor{red!25}RoC Aggregations&\cellcolor{red!25}Text&\cellcolor{red!25}Top-5 Citations&\cellcolor{red!25}21.4/28.5/23.8/51.7/$52.1^*$&\cellcolor{red!25}-\\
\cline{2-7} 
%&\multirow{3}{*}{AusLaw-embedding}&Full Cases&Text&Top-5 Citations&&-\\
%&&Catchwords&Text&Top-5 Citations&&-\\
&AusLaw-embedding&\cellcolor{orange!25}RoC Aggregations&\cellcolor{orange!25}Text&\cellcolor{orange!25}Top-5 Citations&\cellcolor{orange!25}21.2/28.3/23.1/50.8/51.4&\cellcolor{orange!25}-\\
\hlinewd{1pt}
\end{tabular}
}
\caption{Citation accuracy under various methods, including ours. For results marked by $\ddagger$, we used \texttt{gpt-4o-mini} due to cost. Best result is marked with \textbf{$^\ast$}. The rows are colour-coded for readability and easier comparison of settings referenced in the discussion \S\ref{sec:experiment-main}. Text denotes ($S^{i-1}_{c_k}, \texttt{Mask}(S^i_{c_k})$), see \S\ref{subsec:llm} for notations.}
\label{tab:main}
\end{table*}

%\paragraph{Experimental Setups.} 
LoRA~\citep{DBLP:conf/iclr/HuSWALWWC22} and 8-bit quantization was used for the instruction tuning stage. For details regarding the instruction fine-tuning step please refer to Table~\ref{table:fine-tuning-details} of Appendix~\ref{app:instructiontuning}. As evaluation metrics, Accuracy@1 and Accuracy@5 are used. 

\subsection{Results}\label{sec:experiment-main}
The main results are presented in Table~\ref{tab:main}. We structure our discussion of results along the comparisons outlined in the introduction (\S\ref{sec:introduction}).

\paragraph{Pre-training on general text vs. law-specialised text.} The results of the LLM-only experiments~(1st panel of Table~\ref{tab:main}) reveal that pre-training alone is insufficient for achieving satisfactory performance in the citation prediction task. Neither general-domain nor law-specialised models demonstrated reasonable accuracy. For example, Claude Sonnet 3.5 achieved only 15\% accuracy in predicting citations when provided with both the query text and the RoC. In contrast, both the 7B and 54B variants of SaulLM performed even worse, achieving 2\% accuracy or less, despite being explicitly pre-trained on the Open Australian Legal Corpus~(including the NSW Caselaw subset used in this study) and other law-specific datasets. This highlights that mere domain-specific pre-training of relatively large (i.e., SaulLM-54B) is not sufficient for fine-grained tasks. 

Also it is noteworthy that out of the 94-520B tokens used in pre-training of such Lawps-specialised LLMs, only 0.5B tokens were covering Australian jurisdiction (see Table 1s of \citet{saul7b} and \citet{saul54b}). This opens a natural question: whether specific pre-training solely on that segment of 0.5B tokens will give a more jurisdictional chance to the LLM in our task. While a proper investigation of this will require a substantial GPU support, we investigate this question for the 7B-8B LLM scales in Subsection~\ref{sec:auslawllm}.

\paragraph{Pre-training vs. Instruction tuning.} A significant performance boost is observed when comparing pre-trained LLMs to their instruction-tuned counterparts on our training data. For example, the SaulLM-7B model's performance jumps from 0\% to 51.7\% in the more challenging Text-only query setup. Similarly, LLaMA-3.1-8B, a general-purpose model, achieves 46.2\% accuracy after instruction fine-tuning, surpassing its much larger pre-trained 70B counterpart. Notably, between the LLaMA and SaulLM backbones, instruction tuning on the same dataset and for the same number of epochs proves more effective for the domain-specific SaulLM. This underscores the critical importance of not only domain-specialised pre-training but also targeted fine-tuning on task-specific data and requirements. 

\begin{figure}[t] 
    \centering
        \includegraphics[trim={0cm 0.4cm 0 0cm},clip, width=0.5\textwidth]{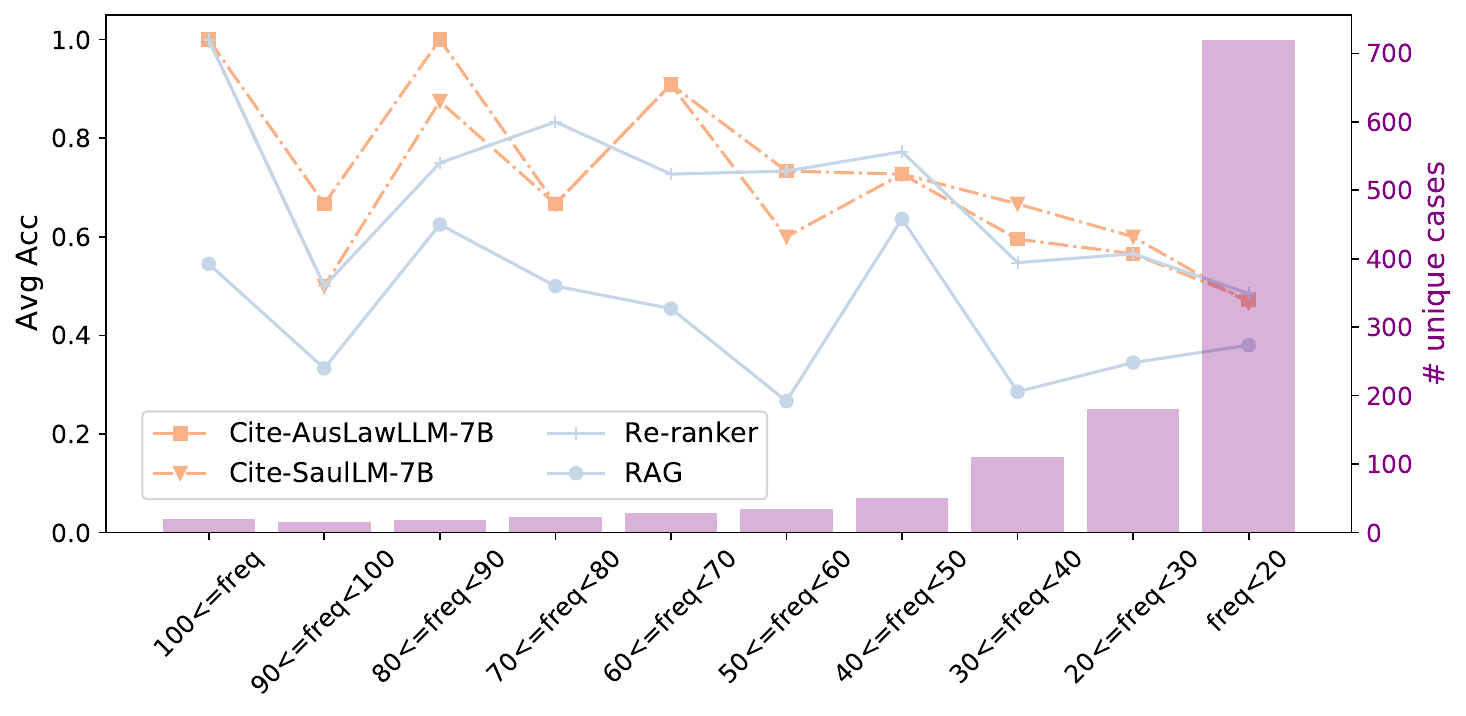}
    \caption{Average Accuracy and Number of Unique Cases for various citation frequency buckets. Accuracies per bin are based on the following settings from Table~\ref{tab:main}: Cite-SaulLM-7B (ACC@1: 51.7),  RAG (ACC@1: 42.9), and Re-ranker (ACC@1: 52.1), and from Table~\ref{tab:pre-trainings}: Cite-AusLawLLM-7B (ACC@1: 52.0).}
    \label{fig:accuracyvsfreq}
\end{figure}

As anticipated, Figure~\ref{fig:accuracyvsfreq} demonstrates a negative correlation between predictive accuracy and the citation frequency in the dataset. Cases cited more than 100 times achieve 100\% accuracy, while accuracy drops below 40\% for cases cited 20 times or fewer. This reflects the challenge fine-tuning faces in accurately predicting less frequently cited cases (i.e., freq<20), where alternative approaches which integrate retrieval (i.e., retrieval with re-ranking, explored next) might offer a more reliable solution.
\paragraph{Generic vs. domain-specialised embeddings.} When comparing the best results of each block that involves retrieval: \colorbox{red!25}{ red rows} (corresponding to the generic text-embedding from OpenAI) vs. \colorbox{orange!25}{orange rows} (corresponding to embeddings trained exclusively on the Open Australian Legal Corpus) in each block of results in Table~\ref{tab:main}, it is important to to acknowledge that this comparison is not entirely fair due to differences in embedding dimensionality (3072 vs. 384), the volume of data used, and the training algorithms employed. Nonetheless, we observe an overall intuitive pattern in favour of the domain-specialised AusLaw-embeddings, or at least being on-par with OpenAIs emeddings (while being $8\times$ smaller in dimensionality). This highlights the benefits of domain-specialised embeddings and suggests a few promising future directions to explore: training larger embeddings tailored to the Australian legal domain, while we do not investigate this further, a trivial direction of exploration could be to build higher dimensional representations, as well as improving the geometric utilisations of the representation space (e.g., through isotropy~\cite{mirrorBert}). We leave further explorations of these to future work.

\paragraph{Database granularity matters.} Probing the results along the \emph{Index Granularity} dimension indicates the significant role the  granularity of database (index) plays in the accuracy of the retrieval system. Contrary to our expectations, Catchwords proved to be the least effective granularity, performing worse than Full Cases. By a substantial margin, the best performance was achieved with the RoC Aggregations granularity. This pattern is consistent for both the generic and law-specialised embeddings, and across all  setups except for the Retrieval-only were Full Case leads to slightly better results compared with Catchwords. 

%% An interesting t-sne experiment could be to show the "transformation of the representation space pre- and post-query expansion" to show what role the LLM plays in re-adjusting the semantic space.

\paragraph{Query expansion vs. voting ensemble vs. RAG.} Comparing all retrieval-based methods, \underline{excluding re-ranking}, hybrid methods are consistently better than Retrieval-only. The \textbf{Voting Ensemble} is the best, followed by \textbf{RAG} and then \textbf{Query Expansion}. The superiority of Voting Ensemble highlights the advantages of combining the predictive quality of instruction-tuned LLMs with the robustness of a retrieval system. In contrast, RAG relies on query augmentation to guide the LLM's predictions within the context of the Top-5 retrieved citations, while the Query Expansion method instead focuses on re-adjusting the query's semantic space before searching the retrieval space.

\paragraph{What we can learn from ACC@5.} 
%When excluding the re-ranking methods, the best result across all settings is achieved via Cite-SaulLM-7B model (i.e., the first row highlighted in green). The second best results are under the Cite-LLaMA-3.1-8B model (both in the LLM-only and the Voting Ensemble setups). 
The general pattern across all experiments suggests the potential that lies within Top-5 retrieved citations. For instance, while the Query expansion results are not competitive at ACC@1, the promising 60.4\% for Cite-SaulLM-7B for ACC@5 (the second row highlighted in red) suggests that with improved re-ranking of the Top-5 hits, there is potential to boost accuracy by up to 10\%. We explore this next.

\paragraph{Re-ranker: the Gap of ACC@1 and ACC@5.} The ACC@5 results indicate that the correct citation is often present within the top five retrieved candidates. Building on this insight, the re-ranker’s goal is to enhance the likelihood of selecting the correct citation from these Top-5 candidates. The bottom two blocks in Table~\ref{tab:main} demonstrate the performance of various re-ranking methods. Notably, the best-performing re-ranker boosts ACC@1 from 35.1\% to 52.1\%, significantly narrowing the gap toward the upper bound defined by ACC@5 (60.4\%).

\subsection{Australian Law Pre-training}\label{sec:auslawllm}
SaulLM-7B was pre-trained on a 94B-token general-domain legal corpus comprising various legal sources, including 0.5B tokens from the Open Australian Legal Corpus. To investigate whether pre-training solely on Australian legal data is sufficient for building an effective Australia-specific legal LLM, we pre-train the same underlying vanilla Mistral-7B model and a LLaMA-3.1-8B model (used in SaulLM), on the 0.5B tokens of Australian Legal corpus. Following pre-training, we apply the same citation instruction-tuning procedure on the citation prediction task. While we only pre-train the underlying LLM for 5 epochs on the 0.5B tokens, due to infrastructure limitations, we observe promising benefits of jurisdictional pre-training. The results, presented in Table~\ref{tab:pre-trainings}, show that Cite-AusLawLLM-7B slightly outperforms Cite-SaulLM-7B (both built on Mistral-7B), despite the latter being trained on a much larger 94B-token corpus. Furthermore, we observe that pre-training significantly boosts performance of the LLaMA-3.1-8B model after Australian Law pre-training pre-training (Cite-AusLawLLM-8B). 
%These findings demonstrate that targeted pre-training on Australian legal data is not only sufficient as pre-training but more effective for downstream citation prediction tasks in the Australian legal domain. 

\begin{table}[t]
\centering
\resizebox{0.48\textwidth}{!}{%
\begin{tabular}{@{}lcc@{}}
\toprule
\textbf{Model} & \textbf{Law Data} &  \textbf{Accuracy} \\ \midrule
Cite-SaulLM-7B (Mixed Law Pre-training)  & 94B                                                    & 51.7                                     \\
Cite-AusLawLLM-7B (AusLaw Pre-training)     & 0.5B                                                & 52.0                                     \\\hline
Cite-LLaMA-3.1-8B (No Law Pre-training)    & -                                                     & 46.2                                     \\
Cite-AusLawLLM-8B (AusLaw Pre-training)   & 0.5B                                                  & 50.0                                     \\ \bottomrule
\end{tabular}}
\caption{The results of different pre-training data (Mixed vs. AusLaw-specific) before doing instruction-tuning on citation prediction task. Data size is reported in tokens.}
\label{tab:pre-trainings}
% \vspace{-4mm}
\end{table}

\section{Conclusion}
In this paper, we proposed a new benchmark, \emph{AusLaw Citation Benchmark}, for legal citation prediction and examined various approaches of leveraging Large Language Models and retrieval systems, evaluating their effectiveness both independently and in combination. Our findings demonstrate that while pre-training large language models on general or even domain-specialised legal texts is a necessary starting point, it is far from sufficient for achieving satisfactory citation prediction accuracy in the Australian legal domain. The most contributing factor for improving performance lies in targeted instruction tuning with task-specific data, which dramatically boosts accuracy. Our experiments further reveal the importance of choosing the right embedding model and database granularity for retrieval, with results showing up to 70\% variation in performance under different granularities. Among the retrieval-augmented methods, ensemble voting strategy stands out as the most effective, outperforming methods like RAG and query expansion. Furthermore, we show that training re-rankers effectively harnesses the untapped gains available in top-k retrieval accuracy. 

While the above approaches offer varying degree of effectiveness, there remains a noticeable $50\%$ gap in performance that calls for further developments. We hope the proposed benchmark and the investigated approaches provide a reliable framework for development of more advanced solutions that intersect LLMs, fine-tuning, pre-training, and retrieval mechanisms.

%As potential for future explorations, training better embeddings (i.e., larger dimensionalities, with specific attention to discriminative features) is a promising direction. Additionally, in this paper we did not test the methods in the out-of-distribution setting which could further highlight other potential challenges in legal domain. 

%Out of distribution was not tested

%Embeddings could be trained

\section*{Limitations}
First, due to computational constraints, we limit most of our open-source experiments to 7B-parameter models and do not evaluate the performance of larger-scale models (i.e., SaulLM 141B), which may offer additional gains. Second, our citation prediction task is currently framed as a single-citation prediction problem, where each input maps to only one citation. A more realistic and challenging setting would involve predicting multiple citations per instance under a larger context. Additionally, in this paper we did not test the methods in the out-of-distribution setting which could further highlight other potential challenges in legal domain. 

\bibliography{acl_latex}

\begin{thebibliography}{34}
\providecommand{\natexlab}[1]{#1}

\bibitem[{Achiam et~al.(2023)Achiam, Adler, Agarwal, Ahmad, Akkaya, Aleman, Almeida, Altenschmidt, Altman, Anadkat et~al.}]{gpt4}
Josh Achiam, Steven Adler, Sandhini Agarwal, Lama Ahmad, Ilge Akkaya, Florencia~Leoni Aleman, Diogo Almeida, Janko Altenschmidt, Sam Altman, Shyamal Anadkat, and 1 others. 2023.
\newblock Gpt-4 technical report.
\newblock \emph{arXiv preprint arXiv:2303.08774}.

\bibitem[{Anthropic(2024)}]{anthropic2024claude35sonnet}
Anthropic. 2024.
\newblock \href {https://www.anthropic.com/news/claude-3-5-sonnet} {Claude 3.5 sonnet}.

\bibitem[{Axel-Lute(1982)}]{AxelLute1982}
Paul Axel-Lute. 1982.
\newblock Legal citation form: Theory and practice.
\newblock \emph{Law Library Journal}, 75:1--27.

\bibitem[{Blair-Stanek et~al.(2023)Blair-Stanek, Holzenberger, and Van~Durme}]{blair2023blt}
Andrew Blair-Stanek, Nils Holzenberger, and Benjamin Van~Durme. 2023.
\newblock Blt: Can large language models handle basic legal text?
\newblock \emph{arXiv preprint arXiv:2311.09693}.

\bibitem[{Chalkidis et~al.(2020)Chalkidis, Fergadiotis, Malakasiotis, Aletras, and Androutsopoulos}]{legalbert}
Ilias Chalkidis, Manos Fergadiotis, Prodromos Malakasiotis, Nikolaos Aletras, and Ion Androutsopoulos. 2020.
\newblock \href {https://doi.org/10.18653/v1/2020.findings-emnlp.261} {{LEGAL}-{BERT}: The muppets straight out of law school}.
\newblock In \emph{Findings of the Association for Computational Linguistics: EMNLP 2020}, pages 2898--2904, Online. Association for Computational Linguistics.

\bibitem[{Chalkidis et~al.(2022)Chalkidis, Jana, Hartung, Bommarito, Androutsopoulos, Katz, and Aletras}]{chalkidis-etal-2022-lexglue}
Ilias Chalkidis, Abhik Jana, Dirk Hartung, Michael Bommarito, Ion Androutsopoulos, Daniel Katz, and Nikolaos Aletras. 2022.
\newblock \href {https://doi.org/10.18653/v1/2022.acl-long.297} {{L}ex{GLUE}: A benchmark dataset for legal language understanding in {E}nglish}.
\newblock In \emph{Proceedings of the 60th Annual Meeting of the Association for Computational Linguistics (Volume 1: Long Papers)}, pages 4310--4330, Dublin, Ireland. Association for Computational Linguistics.

\bibitem[{Cohere(2024)}]{commandrplus}
Cohere. 2024.
\newblock \href {https://docs.cohere.com/v2/docs/command-r-plus} {Command r+}.

\bibitem[{Colombo et~al.(2024{\natexlab{a}})Colombo, Pires, Boudiaf, de~Melo, Hautreux, Malaboeuf, Charpentier, Culver, and Desa}]{saul54b}
Pierre Colombo, Telmo Pires, Malik Boudiaf, Rui Filipe Coimbra~Pereira de~Melo, Gabriel Hautreux, Etienne Malaboeuf, Johanne Charpentier, Dominic Culver, and Michael Desa. 2024{\natexlab{a}}.
\newblock \href {https://openreview.net/forum?id=NLUYZ4ZqNq} {Saul{LM}-54b \& saul{LM}-141b: Scaling up domain adaptation for the legal domain}.
\newblock In \emph{The Thirty-eighth Annual Conference on Neural Information Processing Systems}.

\bibitem[{Colombo et~al.(2024{\natexlab{b}})Colombo, Pires, Boudiaf, Culver, Melo, Corro, Martins, Esposito, Raposo, Morgado, and Desa}]{saul7b}
Pierre Colombo, Telmo~Pessoa Pires, Malik Boudiaf, Dominic Culver, Rui Melo, Caio Corro, Andr{\'{e}} F.~T. Martins, Fabrizio Esposito, Vera~L{\'{u}}cia Raposo, Sofia Morgado, and Michael Desa. 2024{\natexlab{b}}.
\newblock \href {https://doi.org/10.48550/ARXIV.2403.03883} {Saullm-7b: {A} pioneering large language model for law}.
\newblock \emph{CoRR}, abs/2403.03883.

\bibitem[{Dahl et~al.(2024)Dahl, Magesh, Suzgun, and Ho}]{dahl2024large}
Matthew Dahl, Varun Magesh, Mirac Suzgun, and Daniel~E Ho. 2024.
\newblock Large legal fictions: Profiling legal hallucinations in large language models.
\newblock \emph{Journal of Legal Analysis}, 16(1):64--93.

\bibitem[{Deroy et~al.(2024)Deroy, Ghosh, and Ghosh}]{deroy2024applicability}
Aniket Deroy, Kripabandhu Ghosh, and Saptarshi Ghosh. 2024.
\newblock Applicability of large language models and generative models for legal case judgement summarization.
\newblock \emph{Artificial Intelligence and Law}, pages 1--44.

\bibitem[{Dominguez-Olmedo et~al.(2024)Dominguez-Olmedo, Nanda, Abebe, Bechtold, Engel, Frankenreiter, Gummadi, Hardt, and Livermore}]{lawma}
Ricardo Dominguez-Olmedo, Vedant Nanda, Rediet Abebe, Stefan Bechtold, Christoph Engel, Jens Frankenreiter, Krishna Gummadi, Moritz Hardt, and Michael Livermore. 2024.
\newblock Lawma: The power of specialization for legal tasks.
\newblock \emph{arXiv preprint arXiv:2407.16615}.

\bibitem[{Fei et~al.(2023)Fei, Shen, Zhu, Zhou, Han, Zhang, Chen, Shen, and Ge}]{fei2023lawbench}
Zhiwei Fei, Xiaoyu Shen, Dawei Zhu, Fengzhe Zhou, Zhuo Han, Songyang Zhang, Kai Chen, Zongwen Shen, and Jidong Ge. 2023.
\newblock Lawbench: Benchmarking legal knowledge of large language models.
\newblock \emph{arXiv preprint arXiv:2309.16289}.

\bibitem[{Grattafiori et~al.(2024)}]{grattafiori2024llama3}
Aaron Grattafiori and 1 others. 2024.
\newblock \href {https://arxiv.org/abs/2407.21783} {The llama 3 herd of models}.
\newblock \emph{arXiv preprint arXiv:2407.21783}.

\bibitem[{Guha et~al.(2024)Guha, Nyarko, Ho, R{\'e}, Chilton, Chohlas-Wood, Peters, Waldon, Rockmore, Zambrano et~al.}]{guha2024legalbench}
Neel Guha, Julian Nyarko, Daniel Ho, Christopher R{\'e}, Adam Chilton, Alex Chohlas-Wood, Austin Peters, Brandon Waldon, Daniel Rockmore, Diego Zambrano, and 1 others. 2024.
\newblock Legalbench: A collaboratively built benchmark for measuring legal reasoning in large language models.
\newblock \emph{Advances in Neural Information Processing Systems}, 36.

\bibitem[{Hendrycks et~al.(2021)Hendrycks, Burns, Chen, and Ball}]{hendrycks2021cuad}
Dan Hendrycks, Collin Burns, Anya Chen, and Spencer Ball. 2021.
\newblock Cuad: An expert-annotated nlp dataset for legal contract review.
\newblock \emph{arXiv preprint arXiv:2103.06268}.

\bibitem[{Holzenberger et~al.(2020)Holzenberger, Blair{-}Stanek, and Durme}]{DBLP:journals/corr/abs-2005-05257}
Nils Holzenberger, Andrew Blair{-}Stanek, and Benjamin~Van Durme. 2020.
\newblock \href {https://arxiv.org/abs/2005.05257} {A dataset for statutory reasoning in tax law entailment and question answering}.
\newblock \emph{CoRR}, abs/2005.05257.

\bibitem[{Hou et~al.(2024{\natexlab{a}})Hou, Jurayj, Holzenberger, Blair-Stanek, and Van~Durme}]{hou2024gaps}
Abe Hou, William Jurayj, Nils Holzenberger, Andrew Blair-Stanek, and Benjamin Van~Durme. 2024{\natexlab{a}}.
\newblock Gaps or hallucinations? scrutinizing machine-generated legal analysis for fine-grained text evaluations.
\newblock In \emph{Proceedings of the Natural Legal Language Processing Workshop 2024}, pages 280--302.

\bibitem[{Hou et~al.(2024{\natexlab{b}})Hou, Weller, Qin, Yang, Lawrie, Holzenberger, Blair-Stanek, and Durme}]{abe2024clerc}
Abe~Bohan Hou, Orion Weller, Guanghui Qin, Eugene Yang, Dawn Lawrie, Nils Holzenberger, Andrew Blair-Stanek, and Benjamin~Van Durme. 2024{\natexlab{b}}.
\newblock \href {https://arxiv.org/pdf/2406.17186} {Clerc: A dataset for legal case retrieval and retrieval-augmented analysis generation}.
\newblock \emph{ArXiv}.

\bibitem[{Hu et~al.(2022)Hu, Shen, Wallis, Allen{-}Zhu, Li, Wang, Wang, and Chen}]{DBLP:conf/iclr/HuSWALWWC22}
Edward~J. Hu, Yelong Shen, Phillip Wallis, Zeyuan Allen{-}Zhu, Yuanzhi Li, Shean Wang, Lu~Wang, and Weizhu Chen. 2022.
\newblock \href {https://openreview.net/forum?id=nZeVKeeFYf9} {Lora: Low-rank adaptation of large language models}.
\newblock In \emph{The Tenth International Conference on Learning Representations, {ICLR} 2022, Virtual Event, April 25-29, 2022}. OpenReview.net.

\bibitem[{Hu et~al.(2025)Hu, Gan, Xiao, Kuang, and Wu}]{hu2025fine}
Yinghao Hu, Leilei Gan, Wenyi Xiao, Kun Kuang, and Fei Wu. 2025.
\newblock Fine-tuning large language models for improving factuality in legal question answering.
\newblock \emph{arXiv preprint arXiv:2501.06521}.

\bibitem[{Jiang and Yang(2023)}]{10.1145/3594536.3595170}
Cong Jiang and Xiaolei Yang. 2023.
\newblock \href {https://doi.org/10.1145/3594536.3595170} {Legal syllogism prompting: Teaching large language models for legal judgment prediction}.
\newblock In \emph{Proceedings of the Nineteenth International Conference on Artificial Intelligence and Law}, ICAIL '23, page 417–421, New York, NY, USA. Association for Computing Machinery.

\bibitem[{Joshi et~al.(2024)Joshi, Paul, Sharma, Goyal, Ghosh, and Modi}]{indian-bench}
Abhinav Joshi, Shounak Paul, Akshat Sharma, Pawan Goyal, Saptarshi Ghosh, and Ashutosh Modi. 2024.
\newblock Il-tur: Benchmark for indian legal text understanding and reasoning.
\newblock \emph{arXiv preprint arXiv:2407.05399}.

\bibitem[{Li et~al.(2024{\natexlab{a}})Li, Chen, Yang, Ai, Jia, Liu, Lin, Wu, Yuan, Hu et~al.}]{li2024legalagentbench}
Haitao Li, Junjie Chen, Jingli Yang, Qingyao Ai, Wei Jia, Youfeng Liu, Kai Lin, Yueyue Wu, Guozhi Yuan, Yiran Hu, and 1 others. 2024{\natexlab{a}}.
\newblock Legalagentbench: Evaluating llm agents in legal domain.
\newblock \emph{arXiv preprint arXiv:2412.17259}.

\bibitem[{Li et~al.(2024{\natexlab{b}})Li, Chen, Ai, Wu, Zhang, and Liu}]{chinese-bench}
Haitao Li, You Chen, Qingyao Ai, Yueyue Wu, Ruizhe Zhang, and Yiqun Liu. 2024{\natexlab{b}}.
\newblock Lexeval: A comprehensive chinese legal benchmark for evaluating large language models.
\newblock \emph{arXiv preprint arXiv:2409.20288}.

\bibitem[{Liu et~al.(2021)Liu, Vuli{\'c}, Korhonen, and Collier}]{mirrorBert}
Fangyu Liu, Ivan Vuli{\'c}, Anna Korhonen, and Nigel Collier. 2021.
\newblock \href {https://doi.org/10.18653/v1/2021.emnlp-main.109} {Fast, effective, and self-supervised: Transforming masked language models into universal lexical and sentence encoders}.
\newblock In \emph{Proceedings of the 2021 Conference on Empirical Methods in Natural Language Processing}, pages 1442--1459, Online and Punta Cana, Dominican Republic. Association for Computational Linguistics.

\bibitem[{Magesh et~al.(2024)Magesh, Surani, Dahl, Suzgun, Manning, and Ho}]{magesh2024hallucination}
Varun Magesh, Faiz Surani, Matthew Dahl, Mirac Suzgun, Christopher~D Manning, and Daniel~E Ho. 2024.
\newblock Hallucination-free? assessing the reliability of leading ai legal research tools.
\newblock \emph{arXiv preprint arXiv:2405.20362}.

\bibitem[{OpenAI(2024)}]{gpt4o}
OpenAI. 2024.
\newblock \href {https://openai.com/index/hello-gpt-4o/} {Gpt-4o}.

\bibitem[{Peng and Chen(2024)}]{peng2024athena}
Xiao Peng and Liang Chen. 2024.
\newblock Athena: Retrieval-augmented legal judgment prediction with large language models.
\newblock \emph{arXiv preprint arXiv:2410.11195}.

\bibitem[{Pipitone and Alami(2024)}]{legal-rag}
Nicholas Pipitone and Ghita~Houir Alami. 2024.
\newblock Legalbench-rag: A benchmark for retrieval-augmented generation in the legal domain.
\newblock \emph{arXiv preprint arXiv:2408.10343}.

\bibitem[{Pont et~al.(2023)Pont, Galli, Loreggia, Pisano, Rovatti, and Sartor}]{pont2023legal}
Thiago~Dal Pont, Federico Galli, Andrea Loreggia, Giuseppe Pisano, Riccardo Rovatti, and Giovanni Sartor. 2023.
\newblock Legal summarisation through llms: The prodigit project.
\newblock \emph{arXiv preprint arXiv:2308.04416}.

\bibitem[{Schauer(1987)}]{Schauer1987}
Frederick Schauer. 1987.
\newblock Precedent.
\newblock \emph{Stanford Law Review}, 39(3):571--605.

\bibitem[{Shu et~al.(2024)Shu, Zhao, Liu, Demeter, Du, and Zhang}]{lawllm}
Dong Shu, Haoran Zhao, Xukun Liu, David Demeter, Mengnan Du, and Yongfeng Zhang. 2024.
\newblock \href {https://doi.org/10.1145/3627673.3680020} {Lawllm: Law large language model for the {US} legal system}.
\newblock In \emph{Proceedings of the 33rd {ACM} International Conference on Information and Knowledge Management, {CIKM} 2024, Boise, ID, USA, October 21-25, 2024}, pages 4882--4889. {ACM}.

\bibitem[{Wang et~al.(2023)Wang, Scardigli, Tang, Chen, Levkin, Chen, Ball, Woodside, Zhang, and Hendrycks}]{wang2023maud}
Steven~H Wang, Antoine Scardigli, Leonard Tang, Wei Chen, Dimitry Levkin, Anya Chen, Spencer Ball, Thomas Woodside, Oliver Zhang, and Dan Hendrycks. 2023.
\newblock Maud: An expert-annotated legal nlp dataset for merger agreement understanding.
\newblock \emph{arXiv preprint arXiv:2301.00876}.

\end{thebibliography}

\newpage
\appendix
\section{Appendix}
\subsection{Three Examples of Catchwords}\label{app:catchwords}

\begin{figure}[!htbp]
    \centering
    \fbox{
        \parbox{1\linewidth}{
        \textbf{Case 1}: CRIME – Appeals – Appeal against conviction – Unreasonable verdict – two counts of sexual offences – where applicant found guilty on one count and acquitted on the other – whether on all of the evidence it was open to the jury to be satisfied of the applicant’s guilt beyond reasonable doubt – discrepancies and inconsistencies in complainant’s evidence – appeal allowed – conviction quashed.\\\\
        
        \textbf{Case 2}: DEVELOPMENT APPEAL – dual occupancy – contentions resolved - covenant – variation of covenant – weight to be given to covenant - view sharing – community objections.\\\\
        
        \textbf{Case 3}: PROCEDURE – application for separate questions under UCPR 28.2 – where plaintiff seeks orders under Part 1C of the Civil Liability Act 2002 to set aside settlement agreements which may otherwise preclude the plaintiff from maintaining the balance of the proceedings – claimed benefit to defendants of plaintiff’s promise to “take no action”– statutory right to commence proceedings subsequently enacted – construction of s 7D(1) of the Civil Liability Act 2002 – overlap in issues and credit for hearing of separate questions and the balance of the proceedings – application for separate questions rejected. 
        }
    }
    \caption{Examples of Catchwords from different cases in NSW Caselaw.}
      \label{fig:sample}
\end{figure}

\subsection{Distribution of Citations}\label{sec:visual}
\begin{figure*}[!htbp]
    \centering
    \begin{subfigure}
        \centering
        \includegraphics[trim={0cm 0 0 0.73cm},clip, width=0.82\textwidth]{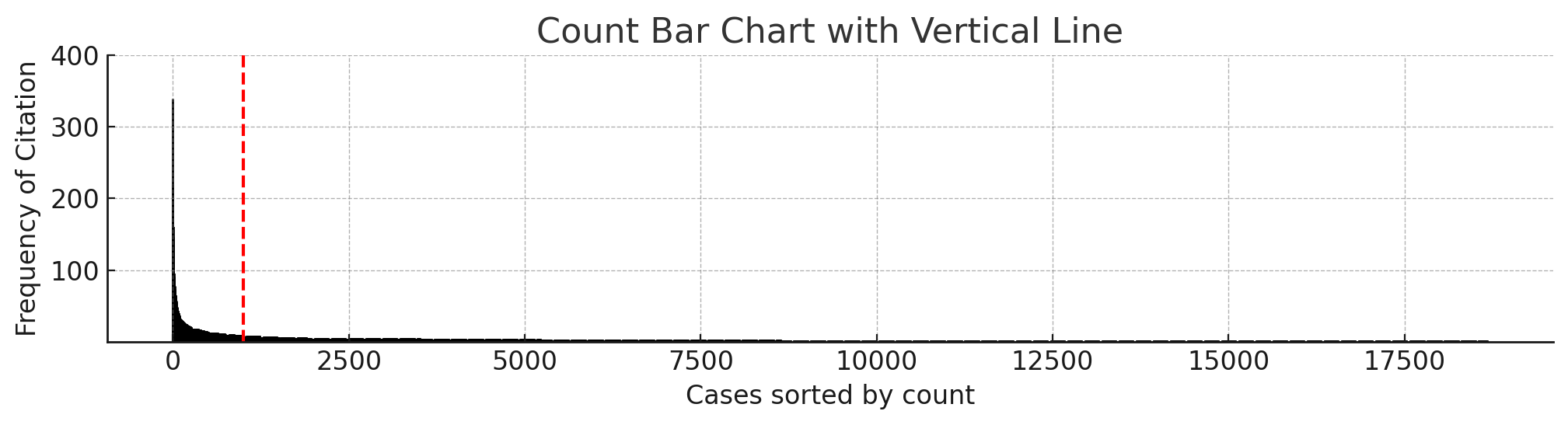}
        %\caption{First instance of the plot}
    \end{subfigure}
    \begin{subfigure}
        \centering
        \includegraphics[trim={0cm 0 1.5cm 0cm},clip, width=0.95\textwidth]{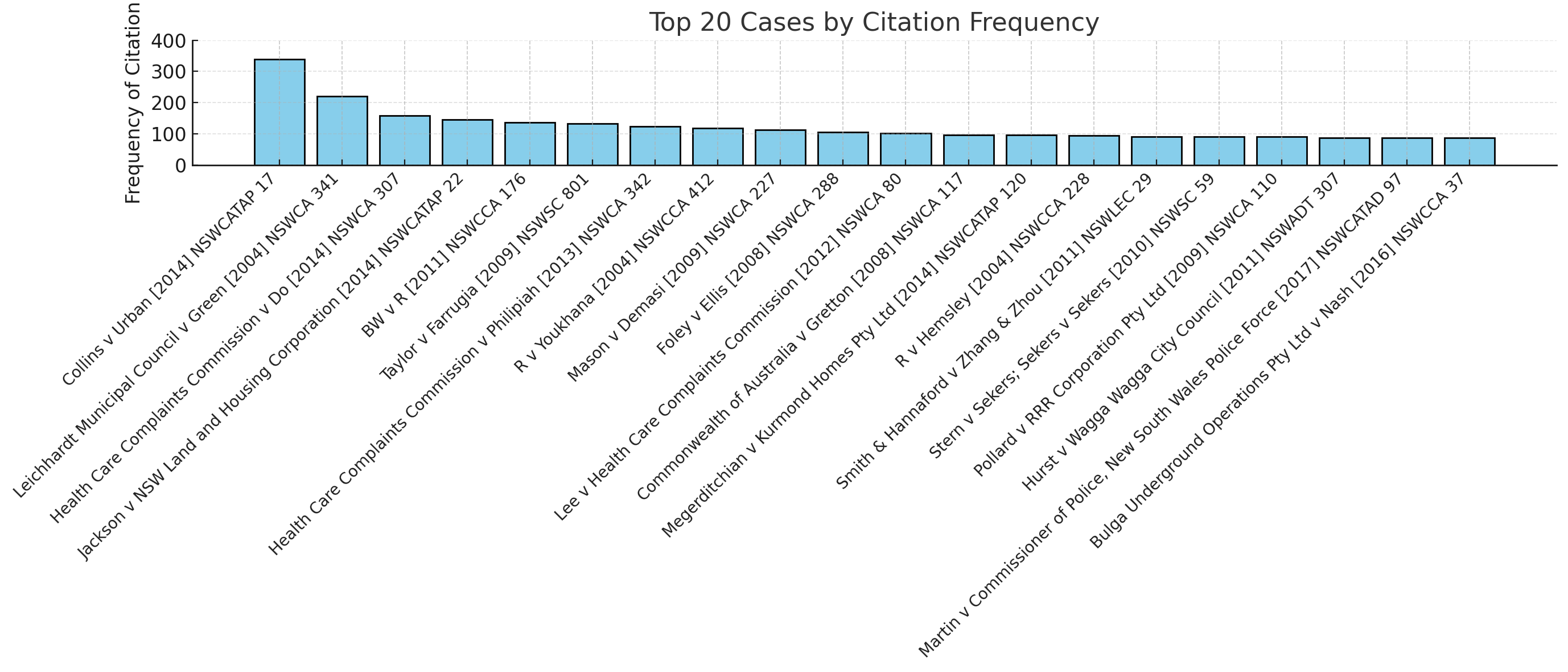}
        %\caption{Second instance of the plot}
    \end{subfigure}
    \caption{(Top) Frequency distribution of unique cases in the data. The red vertical dashed line marks the last case with citation frequency of 9 or higher. (Bottom) Top-20 most frequently cited cases in the data.}
    \label{fig:enter-top20}
\end{figure*}

 Figure~\ref{fig:enter-top20} (Top) shows the frequency distribution of all $18,677$ citations in the data, and Figure~\ref{fig:enter-top20} (Bottom) shows the top-20 most frequent citations. 

\subsection{Details of Instruction Fine-tuning}
The instruction details and training configurations used for instruction fine-tuning are listed in Table~\ref{table:fine-tuning-details} and Table~\ref{reranker_instructions}. The re-rankers use the same training configurations.
\label{app:instructiontuning}

\begin{table*}[!htbp]
\setlength{\extrarowheight}{2pt}
\renewcommand{\arraystretch}{1}
\setlength{\tabcolsep}{5pt}
\resizebox{\linewidth}{!}{%
\begin{tabular}{ll}
\toprule
\textbf{Aspect}             & \textbf{Details} \\ 
\midrule
\textbf{Instruction}        & Predict the name of the case that needs to be cited in the text and explain why it should be cited. \\ 
\textbf{Input}              & \texttt{test\_set[i]['citation\_text'].replace(test\_set[i]['cited\_case\_name'], '<CASENAME>')} \\ 
\textbf{Output}             & \texttt{test\_set[i]['citation\_reason']} + \texttt{<test\_set[i]['citation']>} \\ 
\midrule
\textbf{Training Details}   & \\ 
\quad GPU                   & Single A100 GPU with 80G memory \\ 
\quad Optimizer             & \texttt{adamw\_torch} \\ 
\quad Epochs                & 10 \\ 
\quad Learning Rate         & 2e-4 \\ 
\quad Quantization          & Using 8-bit quantization for LoRA training \\ 
\midrule
\textbf{LoRA Settings}      & \\ 
\quad $r$                   & 16 \\ 
\quad $\alpha$ (LoRA Alpha) & 32 \\ 
\quad Dropout               & 0.05 \\ 
\quad Target Modules        & [up\_proj, down\_proj, gate\_proj, k\_proj, q\_proj, v\_proj, o\_proj] \\ 
\bottomrule
\end{tabular}
}
\caption{Instruction fine-tuning and training details used for training our citation-specialised LLMs.}
\label{table:fine-tuning-details}
\end{table*}

\subsection{Prompt details for various experiments}\label{app:prompt}
All prompting details (or few-shot demonstrations when applicable) used in our experiments are provided in Table~\ref{prompt:roc}-\ref{prompt:citaion_rationale}.

\begin{table*}[t]
\centering
\renewcommand{\arraystretch}{1}
\setlength{\tabcolsep}{5pt}
\begin{tabular}{p{\linewidth}}
\hline
Prompt template for the RoC Generation in our data\\\hline
\begin{lstlisting}[style=text_rm]
system_prompt = """
I have a text from a legal case document that includes a citation to another legal case. A case may be cited for different reasons. You will be provided with the text, the cited case name, and the cited case text.
You must strictly follow these instructions:
1. If the text contains only case names and no further details or context related to the reason for citation, you must generate exactly: NOT ENOUGH INFORMATION.
2. If there is sufficient context in the text to show the reason why the case is cited in this text, you should summarize the reason with a detailed analysis.
3. You should put the reason in the first sentence, and the detailed analysis in the following sentences.
4. You should be conservative as much as possible, do not speculate the reason by yourself even if you may have some knowledge about the case. 
5. Only provide the reason when the text contains enough information to get the reason. Otherwise, just generate NOT ENOUGH INFORMATION.
"""

prompt = """
Text: TEXT
------------------------
Cited Case Name: CASENAME
------------------------
Cited Case Text: CASETEXT
------------------------
Response:
"""
\end{lstlisting}
\\ \hline
\end{tabular}
\caption{The prompt template for the generation of Reason-of-Citation for cases in the data.}\label{prompt:roc}
\end{table*}

\begin{table*}[t]
\centering
\setlength{\extrarowheight}{2pt}
\renewcommand{\arraystretch}{1}
\setlength{\tabcolsep}{5pt}
\begin{tabular}{p{\linewidth}}
\hline
Prompt template used in the LLM-only experiments of Table~\ref{tab:main} \\\hline
\begin{lstlisting}[style=text_rm]
system_prompt = "The following description belongs to a case in the NSW Case Law. You will be given a brief text, and a brief description of a potential citation required. Your task is to predict the citation by listing up-to 5 potential citations, separated by ';'."

prompt = 'Text: ' + test_set[i]['citation_text'].replace(test_set[i]['cited_case_name'], '<CASENAME>') + '\nDescription: ' + test_set[i]['citation_reason'].replace(test_set[i]['citation_reason'], '<CASENAME>')
\end{lstlisting}
\\ \hline
\end{tabular}
\caption{The prompt template used for direct prompting of LLMs (general purpose and law specialised) in Table~\ref{tab:main}. Note, for our instruction-tuned LLMs we used the instructions and not these prompts.} \label{prompt:llmonly}
\end{table*}

\begin{table*}[t]
\centering
\setlength{\extrarowheight}{2pt}
\renewcommand{\arraystretch}{1}
\setlength{\tabcolsep}{5pt}
\begin{tabular}{p{\linewidth}}
\hline
Prompt template and the few-shot demonstrations to generate RoC$^\text{aux}$ in experiments of Table~\ref{tab:main} \\\hline
\begin{lstlisting}[style=text_rm]
system_prompt_few_shot = """
The following text description belongs to a case in the NSW Case Law. You will be given a brief text containing a cited case with a masked token <CASENAME>. Your task is to predict the citation reason of this case. 

Here are some examples for your reference: 

Text: This is not a case where it is appropriate for this Court to deal with the challenges to other findings of breaches of duty by MetLife in its consideration and determination of whether the TPD definition was satisfied. Many of those findings take account of his Honour's conclusion in relation to MetLife's rejection of the lay witness material, and it would be difficult and artificial to deal with those finding on the hypothesis that this rejection involved no breach of duty: see <CASENAME> at [7] (Leeming JA, Basten and Gleeson JJA agreeing).
Citation Reason: The cited case is referenced to support the argument that it would be inappropriate for the Court to address challenges to findings of breaches of duty without considering the context of those findings.

Text: The third order sought was that the Tribunal "award exemplary costs against the Tribunal Registry." As the appellant correctly notes in his submissions the NCAT Act at s 60 sets out that the Tribunal may award costs in proceedings. The Tribunal has found on various occasions that non-parties can be the subject of costs orders: The Owners - Strata Plan No 79749 v Dunstan [2022] NSWCATAP 262; <CASENAME>.
Citation Reason: The cited case is referenced to support the assertion that non-parties can be subject to costs orders in tribunal proceedings.

Text: It does not seem to me that the appellant necessarily intends them in that manner, however that is the legal effect of the applications which are not properly brought before the Tribunal and cannot be determined in the appellant's favour. It is the objectively ascertained purpose of the applications, and not the appellant's subjective intent, which is relevant in that regard: <CASENAME> at [28].
Citation Reason: The cited case is referenced to emphasize the distinction between the subjective intent of the appellant and the objective legal effect of the applications.
"""

prompt = 'Text: ' + test_set[i]['citation_text'].replace(test_set[i]['cited_case_name'], '<CASENAME>')
\end{lstlisting}
\\ \hline
\end{tabular}
\caption{The prompt template used for generating auxiliary RoC (i.e., RoC$^{aux}$) from LLMs in Table~\ref{tab:main}.} \label{prompt:llm-retrieval}
\end{table*}

\begin{table*}[t]
\centering
\setlength{\extrarowheight}{2pt}
\renewcommand{\arraystretch}{1}
\setlength{\tabcolsep}{5pt}
\begin{tabular}{p{\textwidth}}
\hline
The prompt template used for LLMs in the RAG experiment of Table~\ref{tab:main} \\
\hline
\begin{lstlisting}[style=text_rm]
catchwords_rank_sys_prompt = """
The following description belongs to a case in the NSW Case Law, but with a missing citation showing <CASENAME>. You will be given a brief text, 5 potential citations and their corresponding catchwords. Your task is to rank the 5 potential citations according to what is most likely to be the correct citation in the text. Show your ranking result in a list, separated by '\n'.
"""

catchwords_prompt = """
Text:
TEXT

Potential Citations:

CITATION1
Catchwords: CATCHWORDS1

CITATION2
Catchwords: CATCHWORDS2

CITATION3
Catchwords: CATCHWORDS3

CITATION4
Catchwords: CATCHWORDS4

CITATION5
Catchwords: CATCHWORDS5

"""
\end{lstlisting}
\\
\hline
\end{tabular}
\caption{The prompt template used for re-ranking with LLMs in the RAG experiment—corresponding to Catch Words—in Table~\ref{tab:main}. Similar prompts are used for RoC Aggregations and Full Cases experiments.}
\label{prompt:llm-ranking}
\end{table*}

\begin{table*}[t]
\centering
\renewcommand{\arraystretch}{1}
\setlength{\tabcolsep}{5pt}
\begin{tabular}{p{\linewidth}}
\hline
Prompt template for merging the citation reasons of each case in re-ranker setting 2\\\hline
\begin{lstlisting}[style=text_rm]
system_prompt = """
You will be provided with a case from NSW Case Law, along with reasons it may be cited. Your task is to concisely summarize the reasons into a few sentences (2~3 sentences), if the original citation reasons are too long.
"""

prompt = """
Casename:
CASENAME

Citation Reasons: 
CITATIONREASON
"""
\end{lstlisting}
\\ \hline
\end{tabular}
\caption{The prompt template for merging the citation reasons of each case into text of similar length.}\label{prompt:merge}
\end{table*}

\begin{table*}[t]
\centering
\setlength{\extrarowheight}{2pt}
\renewcommand{\arraystretch}{1}
\setlength{\tabcolsep}{5pt}
\begin{tabular}{p{\textwidth}}
\hline
The prompt template used for generating the citation rationale in re-ranker setting 3 \\
\hline
\begin{lstlisting}[style=text_rm]
ranker_reason_sys_prompt = """
The following description belongs to a case in the NSW Case Law, you should predict the citation in the text. 
You will be given a brief text, 5 potential citations, their corresponding citation reasons and a ground-truth correct citation with its potential citation reasons. 
Note that the ground-truth citation may not be included among the five potential options.
Your task is to predict the missing citation, provide a rationale before you draw a conclusion.
Your predicted citation should be the ground-truth correct citation, but don't let on that you already know the answer (Don't mention anything about the ground-truth).
The rationale should not be over 2~3 sentences.
"""

ranker_reason_prompt = """
Text:
TEXT

Potential Citations:

CITATION1
Citation Reasons: CITATIONREASON1

CITATION2
Citation Reasons: CITATIONREASON2

CITATION3
Citation Reasons: CITATIONREASON3

CITATION4
Citation Reasons: CITATIONREASON4

CITATION5
Citation Reasons: CITATIONREASON5

Ground-truth Correct Citation:
GROUNDTRUTH

Ground-truth Citation Reasons:
GTREASON
"""
\end{lstlisting}
\\
\hline
\end{tabular}
\caption{The prompt template used for generating the ground-truth citation rationale given candidate citations in re-ranker setting 3.}
\label{prompt:citaion_rationale}
\end{table*}

\begin{table*}[t]\small
\centering
\resizebox{0.8\textwidth}{!}{%
\begin{tabular}{@{}lll@{}}
\toprule
                                      & \textbf{Instruction} & Predict the citation in the text.                                                                                                                                                                                                                                                                                                                                                                                                                                                                                              \\ \midrule
\multirow{2}{*}{\textbf{Setting 1}}   & \textbf{Input}                & \begin{tabular}[c]{@{}l@{}}\texttt{test\_set[i]['citation\_text'].replace(test\_set[i]['cited\_case\_name'], '<CASENAME>')} + \\ Potential Citations:\\ CITATION1\\ Citation Reasons: FIRST\_CITATIONREASON1\\ CITATION2\\ Citation Reasons: FIRST\_CITATIONREASON2\\ CITATION3\\ Citation Reasons: FIRST\_CITATIONREASON3\\ CITATION4\\ Citation Reasons: FIRST\_CITATIONREASON4\\ CITATION5\\ Citation Reasons: FIRST\_CITATIONREASON5\end{tabular}                  \\ \cmidrule(l){2-3} 
                                      & \textbf{Output}      & \textless{}GOLD CITATION\textgreater{}                                                                                                                                                                                                                                                                                                                                                                                                                                                                                         \\ \midrule
\multirow{2}{*}{\textbf{Setting 2}}   & \textbf{Input}       & \begin{tabular}[c]{@{}l@{}}\textbackslash{}\texttt{test\_set[i]['citation\_text'].replace(test\_set[i]['cited\_case\_name'], '<CASENAME>')} + \\ Potential Citations:\\ CITATION1\\ Citation Reasons: MERGED\_CITATIONREASON1\\ CITATION2\\ Citation Reasons: MERGED\_CITATIONREASON2\\ CITATION3\\ Citation Reasons: MERGED\_CITATIONREASON3\\ CITATION4\\ Citation Reasons: MERGED\_CITATIONREASON4\\ CITATION5\\ Citation Reasons: MERGED\_CITATIONREASON5\end{tabular}          \\ \cmidrule(l){2-3} 
                                      & \textbf{Output}      & GOLD\_CITATION\_REASON + \textless{}GOLD CITATION\textgreater{}                                                                                                                                                                                                                                                                                                                                                                                                                                                                \\ \midrule
\multirow{2}{*}{\textbf{Setting 3}}   & \textbf{Input}       & \begin{tabular}[c]{@{}l@{}}\texttt{test\_set[i]['citation\_text'].replace(test\_set[i]['cited\_case\_name'], '<CASENAME>')} + \\ Potential Citations:\\ CITATION1\\ Citation Reasons: MERGED\_CITATIONREASON1\\ CITATION2\\ Citation Reasons: MERGED\_CITATIONREASON2\\ CITATION3\\ Citation Reasons: MERGED\_CITATIONREASON3\\ CITATION4\\ Citation Reasons: MERGED\_CITATIONREASON4\\ CITATION5\\ Citation Reasons: MERGED\_CITATIONREASON5\end{tabular}          \\ \cmidrule(l){2-3} 
                                      & \textbf{Output}      & RATIONAL + \textless{}GOLD CITATION\textgreater{}                                                                                                                                                                                                                                                                                                                                                                                                                                                                              \\ \midrule
\multirow{2}{*}{\textbf{Setting 4v1}} & \textbf{Input}       & \begin{tabular}[c]{@{}l@{}}\texttt{test\_set[i]['citation\_text'].replace(test\_set[i]['cited\_case\_name'], '<CASENAME>')} +\\ Potential Citations:\\ CITATION1\\ Citation Reasons: Top-1\_CITATIONREASON1\\ CITATION2\\ Citation Reasons: Top-1\_CITATIONREASON2\\ CITATION3\\ Citation Reasons: Top-1\_CITATIONREASON3\\ CITATION4\\ Citation Reasons: Top-1\_CITATIONREASON4\\ CITATION5\\ Citation Reasons: Top-1\_CITATIONREASON5\end{tabular}                \\ \cmidrule(l){2-3} 
                                      & \textbf{Output}      & RETRIEVED\_TOP\_1\_CITATION\_REASON + \textless{}GOLD CITATION\textgreater{}                                                                                                                                                                                                                                                                                                                                                                                                                                                   \\ \midrule
\multirow{2}{*}{\textbf{Setting 4v2}} & \textbf{Input}       & \begin{tabular}[c]{@{}l@{}}\texttt{test\_set[i]['citation\_text'].replace(test\_set[i]['cited\_case\_name'], '<CASENAME>')} + \\ Potential Citations:\\ CITATION1\\ Citation Reasons: Top-1\_CITATIONREASON1\\ CITATION2\\ Citation Reasons: Top-1\_CITATIONREASON2\\ CITATION3\\ Citation Reasons: Top-1\_CITATIONREASON3\\ CITATION4\\ Citation Reasons: Top-1\_CITATIONREASON4\\ CITATION5\\ Citation Reasons: Top-1\_CITATIONREASON5\end{tabular}                \\ \cmidrule(l){2-3}
                                      & \textbf{Output}      & GOLD\_CITATION\_REASON + \textless{}GOLD CITATION\textgreater{}                                                                                                                                                                                                                                                                                                                                                                                                                                                                \\ \bottomrule
\end{tabular}}
\caption{The used instruction, input and output for the training of different re-rankers.}
\label{reranker_instructions}
\end{table*}

\end{document}